%% file: main.tex
\pdfoutput=1
\documentclass{applemlr}

\input{apple_preamble}

\usepackage[utf8]{inputenc}
\usepackage{latexsym}
\usepackage{listings}
\title{Plan or Retrieve First? A Scale Study of Tool Discovery Architectures}

\author{Mingfei Zhang}
\author{Amit Surana}
\author{Ziauddin Majid}
\author{Chenxi Wang}
\author{Michael Xu}

\affiliation{Apple}

\abstract{
Tool-augmented LLM agents discover relevant APIs from large registries before planning execution, yet finite context windows prohibit injecting all tool definitions into a single prompt. The dominant approach, which we term Retrieve-then-Plan (RtP), embeds the user query and retrieves similar tools via approximate nearest-neighbor search. We study an alternative Plan-then-Retrieve (PtR) paradigm in which the LLM first generates Abstract Tool Stubs (hypothesized function signatures describing needed capabilities) which then serve as retrieval queries against the registry. Evaluating on ToolBench (46{,}984 unique API endpoints, 50 categories) across single-tool (G1), intra-category multi-tool (G2), and cross-category multi-tool (G3) query subsets, we find a clear complexity-dependent crossover: RtP outperforms PtR on single-tool queries while PtR outperforms RtP on multi-tool queries. This crossover holds at every retrieval budget ($k{=}1$ to $20$), is robust across registry sizes from 1k to 47k APIs, and is consistent across model families (Gemini 2.5 Flash and gpt-oss-120b). We characterize a stub-hallucination failure mode (the LLM invents plausible but non-existent API names) that acts as a double-edged sword: on simple queries PtR's stubs abstract away proper-noun signals that RtP matches directly, yet on complex queries the same mechanism decomposes multi-faceted intent into functional sub-needs. We further analyze token overhead and show that PtR's extra discovery call roughly doubles inference cost, which pays off on multi-tool queries but hurts on single-tool ones, motivating complexity-adaptive routing as a practical direction — one that can be driven by a zero-cost signal already present in the pipeline (RtP's top-1 retrieval score).
}

\metadata[Venue]{\sffamily Accepted at EMNLP 2026 (Industry Track)}
\metadata[Correspondence]{\sffamily \url{mingfei_zhang@apple.com}}
\date{\sffamily 01 Jun 2026}

\begin{document}

\maketitle

\section{Introduction}

Tool-augmented LLM agents operate over registries of thousands of APIs under tight context constraints~\citep{qin2023toolbench,patil2024gorilla}. The dominant approach directly embeds the user query and retrieves the top-$k$ most similar APIs before passing them to the planner~\citep{patil2024gorilla,qin2023toolbench,yuan2024craft}, assuming the semantic gap between user intent and API documentation is small enough to bridge by embedding alone. Recent work shows this assumption often fails~\citep{shi2025toolret,huang2024metatool}, motivating an alternative direction where the LLM generates query reformulations or plans before retrieval to better express functional intent~\citep{fang2026toolqp}. However, such works propose training-dependent systems aimed at improving retrieval quality, rather than characterizing when and why each approach is preferable. The underlying paradigm comparison remains unexplored.

\begin{figure*}[t]
\centering
\includegraphics[width=0.85\textwidth]{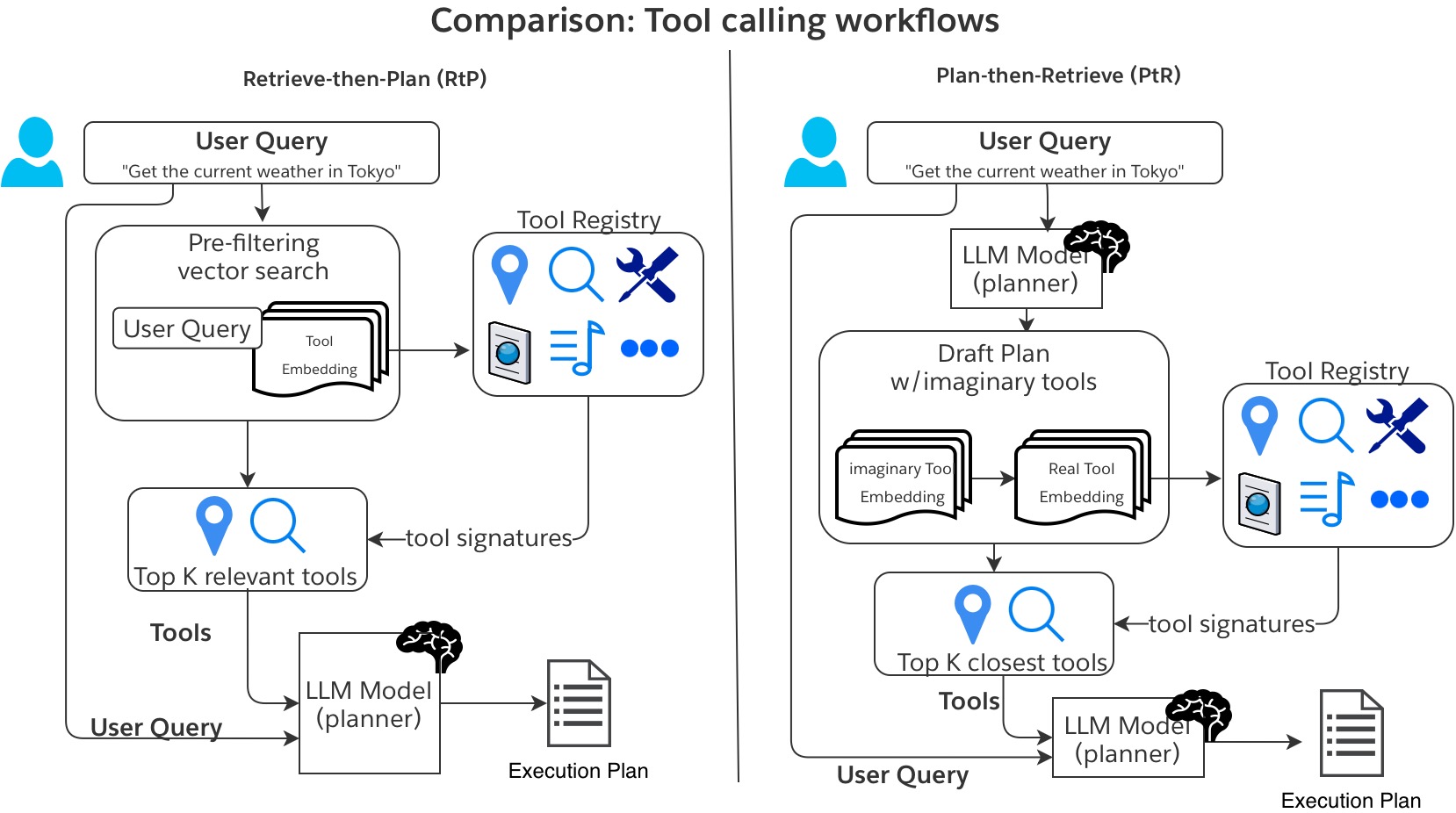}
\caption{Comparison of the two discovery architectures. \textbf{Left (RtP)}: the user query is embedded directly and matched against the tool registry via vector search; the top-$k$ results are passed to the LLM planner. \textbf{Right (PtR)}: the LLM first generates a draft plan with imaginary tool stubs, whose embeddings are used to query the registry; the matched real tools are returned to the planner for final execution planning.}
\label{fig:architecture}
\end{figure*}

Prior work has studied retrieval-side improvements: Tool Retrieval Bridge~\citep{chen2026trb} rewrites queries into retriever-friendly forms; Re-Invoke~\citep{chen2024reinvoke} enriches tool descriptions with synthetic queries; TaskBench~\citep{shen2024taskbench} benchmarks the decompose-then-select pipeline; and Tool-MVR~\citep{ma2025toolmvr} treats stub hallucination as a bug to fix (we show it is a feature on complex queries). ToolFlood~\citep{jawad2026toolflood} and a recent survey~\citep{qin2023toollearning} provide broader context on retrieval robustness and tool learning. A related line of work uses LLM-guided tool selection, where the model draws on parametric knowledge rather than vector retrieval: HuggingGPT~\citep{shen2023hugginggpt} has ChatGPT select specialized models from HuggingFace; AnyTool~\citep{du2024anytool} uses GPT-4 to hierarchically narrow API categories; and LATM~\citep{cai2023large} has LLMs create tools from scratch.

The closest prior work to the paradigm we study is TOOLQP~\citep{fang2026toolqp}, which generates multi-round query reformulations to improve API retrieval, trained with reinforcement learning. Our approach differs in three ways: stubs are generated in a single zero-shot prompt call with no training, we generate hypothetical function signatures rather than query reformulations, and we study PtR as a controlled paradigm rather than proposing a new system. Despite this active area, no prior work has directly compared these two retrieval paradigms in a controlled setting at large registry scale, characterized how the trade-off changes with query complexity, or analyzed how retrieval differences cascade into planner behavior.

We term the dominant paradigm Retrieve-then-Plan (RtP) and the alternative Plan-then-Retrieve (PtR), and present the first systematic apples-to-apples comparison at industrial scale (50k APIs). A key analytical lens we introduce is Intent-Registry Coupling: the semantic gap between how users describe tasks and how APIs are documented. Our PtR instantiation generates Abstract Tool Stubs (hypothesized function signatures) as retrieval queries in a single zero-shot prompt call with no training. Our contributions: \textbf{(1)} a complexity-dependent crossover (RtP wins on simple queries, PtR wins on multi-tool queries) robust to retrieval budget, registry size, and model family; \textbf{(2)} a characterization of stub hallucination as a dual mechanism: it degrades retrieval when queries have high Intent-Registry Coupling, and improves it when they do not; \textbf{(3)} an analysis of token overhead showing that directing single-tool queries to RtP and multi-tool queries to PtR reduces discovery-call overhead by ${\sim}36\%$ with no retrieval quality loss, motivating complexity-adaptive routing as a practical direction consistent with prior work on complexity-aware retrieval~\citep{jeong2024adaptiverag}.

\section{Method}

\begin{figure*}[!t]
\centering
\includegraphics[width=0.8\textwidth]{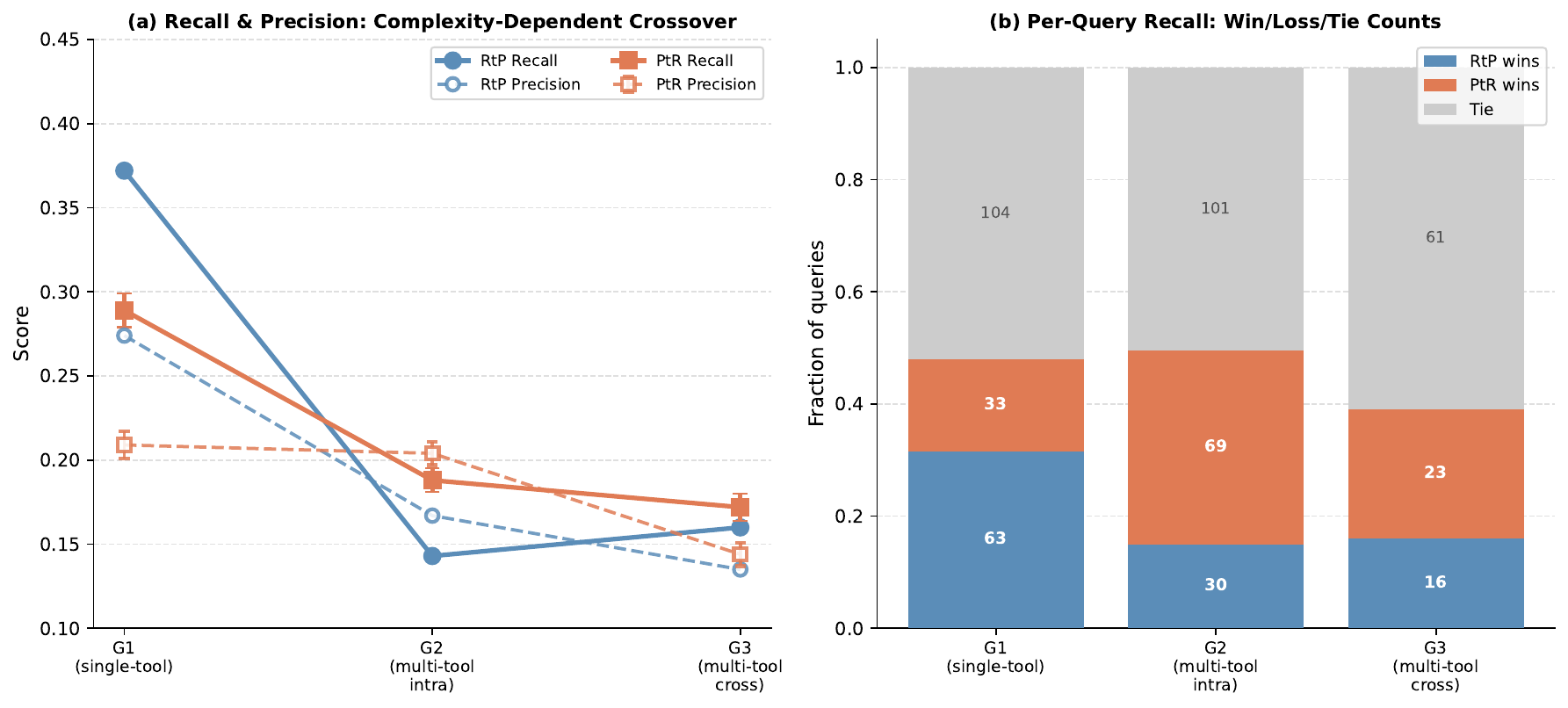}
\caption{Complexity-dependent crossover. (a) Both recall (solid) and precision (dashed) flip from RtP-dominant on G1 to PtR-dominant on G2/G3. (b) Per-query recall win/loss counts: RtP wins more queries on G1 (63 vs.\ 33); PtR wins more on G2 (69 vs.\ 30).}
\label{fig:crossover}
\end{figure*}

\subsection{Architectures}
Figure~\ref{fig:architecture} illustrates both pipelines. Our RtP and PtR are purpose-built, controlled instantiations of these paradigms sharing identical infrastructure. Both conditions share the same infrastructure: GTE-base~\citep{li2023towards} embeddings, a FAISS~\citep{johnson2019billion} flat index over 46{,}984 deduplicated ToolBench API endpoints, and Gemini 2.5 Flash~\citep{googledeepind2025gemini25flash} as both the planner and (for PtR) the discovery model (temperature $= 0$, retrieval budget $k{=}5$). Let $\mathcal{T}$ be the tool registry, $q$ the user query, $d_t$ the documentation of tool $t$, $\mathrm{sim}(\cdot,\cdot)$ the cosine similarity between embeddings, and $\pi(\cdot)$ the agent's execution plan given a retrieved tool set.

\textbf{RtP} embeds the raw query and retrieves the top-$k$ most similar APIs: $R_\text{RtP}(q) = \mathrm{TopK}_{t\in \mathcal{T}}\,\mathrm{sim}(q,d_t)$; the planner then produces plan $\pi_\text{RtP}(q, R_\text{RtP}(q))$.

\textbf{PtR} first sends $q$ to the planner \emph{without} any tool definitions, generating stubs $S(q)=\{s_1,\ldots,s_m\}$ (JSON objects with a hypothesized tool name and reasoning). Each stub is embedded and matched against the index; the union of top-$k$ results (deduplicated, re-ranked) forms $R_\text{PtR}(q) = \mathrm{TopK}_{t\in \mathcal{T}}\max_{s\in S(q)}\mathrm{sim}(s,d_t)$, which is passed back to the planner to produce $\pi_\text{PtR}(q, R_\text{PtR}(q))$.

An unfiltered baseline is infeasible at this scale: ToolBench categories contain up to 4{,}455 APIs (Sports) and 3{,}583 (Data), and Ground-Truth APIs are distributed across each category such that any fixed-size truncation misses them in the vast majority of cases.

\subsection{Data and Metrics}
\label{sec:method}
ToolBench~\citep{qin2023toolbench} provides three test subsets: \textbf{G1} (single-tool, $n{=}200$), \textbf{G2} (intra-category multi-tool, $n{=}200$), and \textbf{G3} (cross-category multi-tool, $n{=}100$; smaller set, so G3 differences have lower statistical power). We report retrieval \textbf{recall} and \textbf{precision} computed at the fixed budget $k{=}5$ against the Ground-Truth API set from each query's ToolBench gold trajectory (all APIs executed in the reference solution). RtP retrieval is deterministic (the query is embedded by GTE-base, a fixed model, and matched against the FAISS index via exact nearest-neighbor search; no LLM call, no sampling), so its precision and recall carry no run-to-run variance and are reported without a standard deviation. All remaining quantities — PtR's precision and recall, and pass rate for both architectures — are averaged over 4 independent runs and reported as mean$\,\pm\,$std, since both stub generation and planner output vary across runs. Statistical comparisons use a one-sample $t$-test for precision and recall (PtR's 4-run mean vs.\ the deterministic RtP value) and a two-sample $t$-test for pass rate. All experiments use the full 46{,}984-API registry unless otherwise noted.

\textit{Evaluation proxy.} We evaluate against ToolBench gold trajectories as a retrieval proxy rather than end-to-end API execution. Trajectory fidelity (at least $96\%$; \S\ref{sec:trajectory}) supports this as a valid proxy for planner behavior: the planner almost always selects from the retrieved set, so retrieval quality upper-bounds task completion. Parameter correctness and final answer quality are not assessed and constitute a limitation noted in \S\ref{sec:limitations}.

\section{Results}

\subsection{Main Results}
Figure~\ref{fig:crossover} presents the central finding (full numerical results in Appendix~\ref{app:main_table}). RtP leads all G1 metrics: recall $0.372$ vs.\ $0.289{\pm}0.010$ ($p{=}0.002$), precision $0.274$ vs.\ $0.209{\pm}0.008$ ($p{<}0.001$), pass rate $0.884{\pm}0.015$ vs.\ $0.840{\pm}0.024$ ($p{=}0.022$). The pattern reverses on multi-tool queries: PtR leads G2 recall $0.188{\pm}0.007$ vs.\ $0.143$ ($p{<}0.001$), precision $0.204{\pm}0.007$ vs.\ $0.167$ ($p{=}0.039$), and pass rate $0.855{\pm}0.020$ vs.\ $0.747{\pm}0.015$ ($p{<}0.001$). On G3 ($n{=}100$), PtR maintains a directional advantage: recall $0.172{\pm}0.008$ vs.\ $0.160$ and pass rate $0.820{\pm}0.014$ vs.\ $0.732{\pm}0.010$ ($p{<}0.001$); the recall difference does not reach significance, reflecting the smaller G3 sample. Panel (b) shows the per-query breakdown: on G1, RtP wins on 63 queries vs.\ PtR's 33; on G2 the pattern flips, with PtR winning on 69 queries vs.\ RtP's 30.

Table~\ref{tab:k_ablation} shows the crossover holds at every retrieval budget from $k{=}1$ to $k{=}20$. PtR's G2 advantage grows with $k$ (from $+0.014$ at $k{=}1$ to $+0.066$ at $k{=}20$) because each stub contributes its own top-$k$ candidates, providing broader multi-faceted coverage as $k$ increases. RtP's G1 advantage is stable across $k$.

\begin{table}[t]
\centering
\small
\caption{Recall at varying retrieval budget $k$. \textbf{Bold}: better architecture. Crossover holds at every $k$. Statistical significance of the RtP--PtR difference at $k{=}5$ is reported in Table~\ref{tab:overall_app}.}
\label{tab:k_ablation}
\begin{tabular}{ccccccc}
\toprule
& \multicolumn{2}{c}{\textbf{G1}} & \multicolumn{2}{c}{\textbf{G2}} & \multicolumn{2}{c}{\textbf{G3}} \\
\cmidrule(lr){2-3}\cmidrule(lr){4-5}\cmidrule(lr){6-7}
$k$ & RtP & PtR & RtP & PtR & RtP & PtR \\
\midrule
1  & $\mathbf{.161}$ & $.127$ & $.065$ & $\mathbf{.079}$ & $.054$ & $\mathbf{.075}$ \\
3  & $\mathbf{.307}$ & $.223$ & $.110$ & $\mathbf{.148}$ & $.105$ & $\mathbf{.139}$ \\
5  & $\mathbf{.372}$ & $.279$ & $.143$ & $\mathbf{.184}$ & $.160$ & $\mathbf{.177}$ \\
10 & $\mathbf{.453}$ & $.356$ & $.186$ & $\mathbf{.234}$ & $.217$ & $\mathbf{.229}$ \\
20 & $\mathbf{.524}$ & $.444$ & $.229$ & $\mathbf{.295}$ & $.288$ & $\mathbf{.299}$ \\
\bottomrule
\end{tabular}
\end{table}

\subsection{Scaling Behavior}
\begin{figure}[t]
\centering
\includegraphics[width=0.6\textwidth]{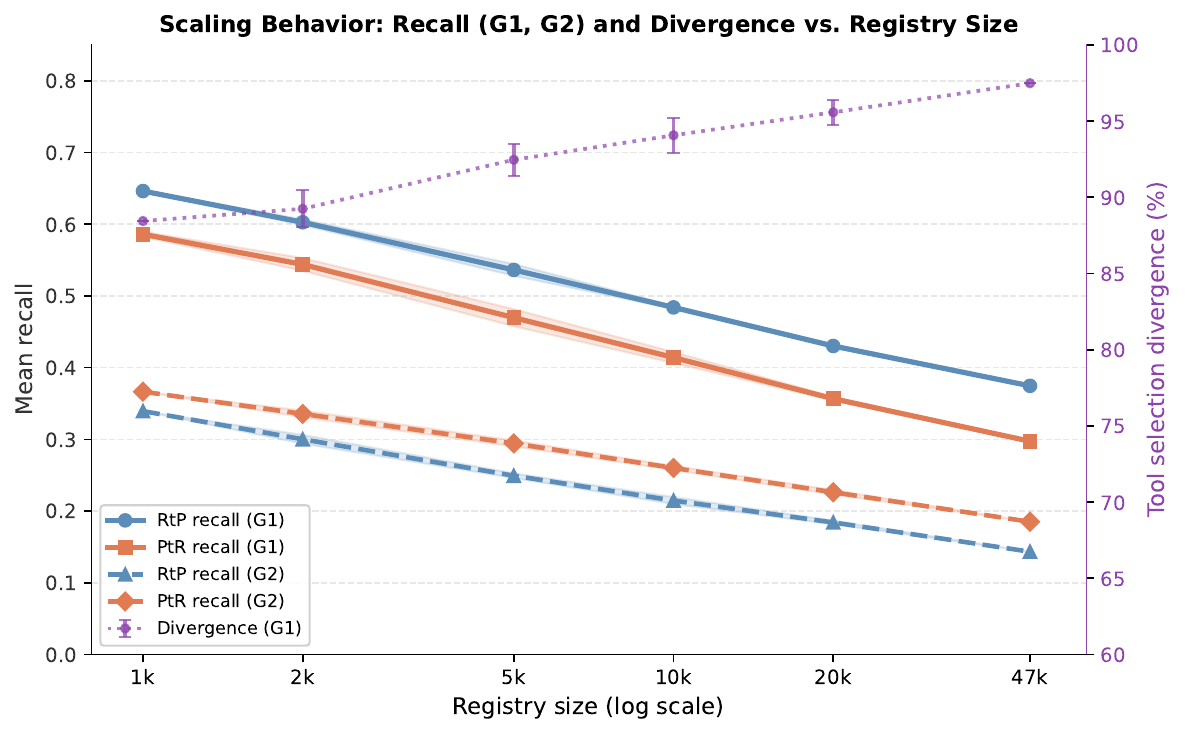}
\caption{Recall and tool selection divergence vs.\ registry size. Shaded bands show $\pm$std across 5 random seeds; the full-registry point (46{,}984 APIs) uses a single deterministic run. G1 (RtP advantage) and G2 (PtR advantage) recall curves are both shown; divergence grows monotonically with registry size.}
\label{fig:scaling}
\end{figure}

Figure~\ref{fig:scaling} shows recall for both G1 and G2, alongside \textbf{tool selection divergence} (fraction of queries where the two architectures retrieve entirely non-overlapping top-$k$ API sets) across sub-registries from 1k to 47k APIs (5 random seeds, proportional category sampling). Both recall curves degrade with registry size but the RtP--PtR gap is stable across scales (recall std $<0.012$): RtP maintains its G1 advantage and PtR maintains its G2 advantage at every registry size tested. Tool selection divergence increases from $88\%$ at 1k to $98\%$ at 47k (Table~\ref{tab:scaling_raw}), confirming the architecture choice is a structural property that becomes more consequential at scale. Per-subset divergence values for the full registry are reported in Appendix~\ref{app:divergence}.

\subsection{Agent Trajectory Divergence}
We define the \textit{agent trajectory} as the set of APIs the planner names and the steps it describes in its output. Since we do not execute API calls, all measurements reflect the planner's \emph{intended} behavior. Trajectory divergence measures whether the two architectures lead to the same intended tool calls (via Jaccard similarity). Trajectory fidelity measures whether the planner stays within the retrieved set (100\% = planner always names an API it was given).

\begin{figure*}[t]
\centering
\includegraphics[width=\textwidth]{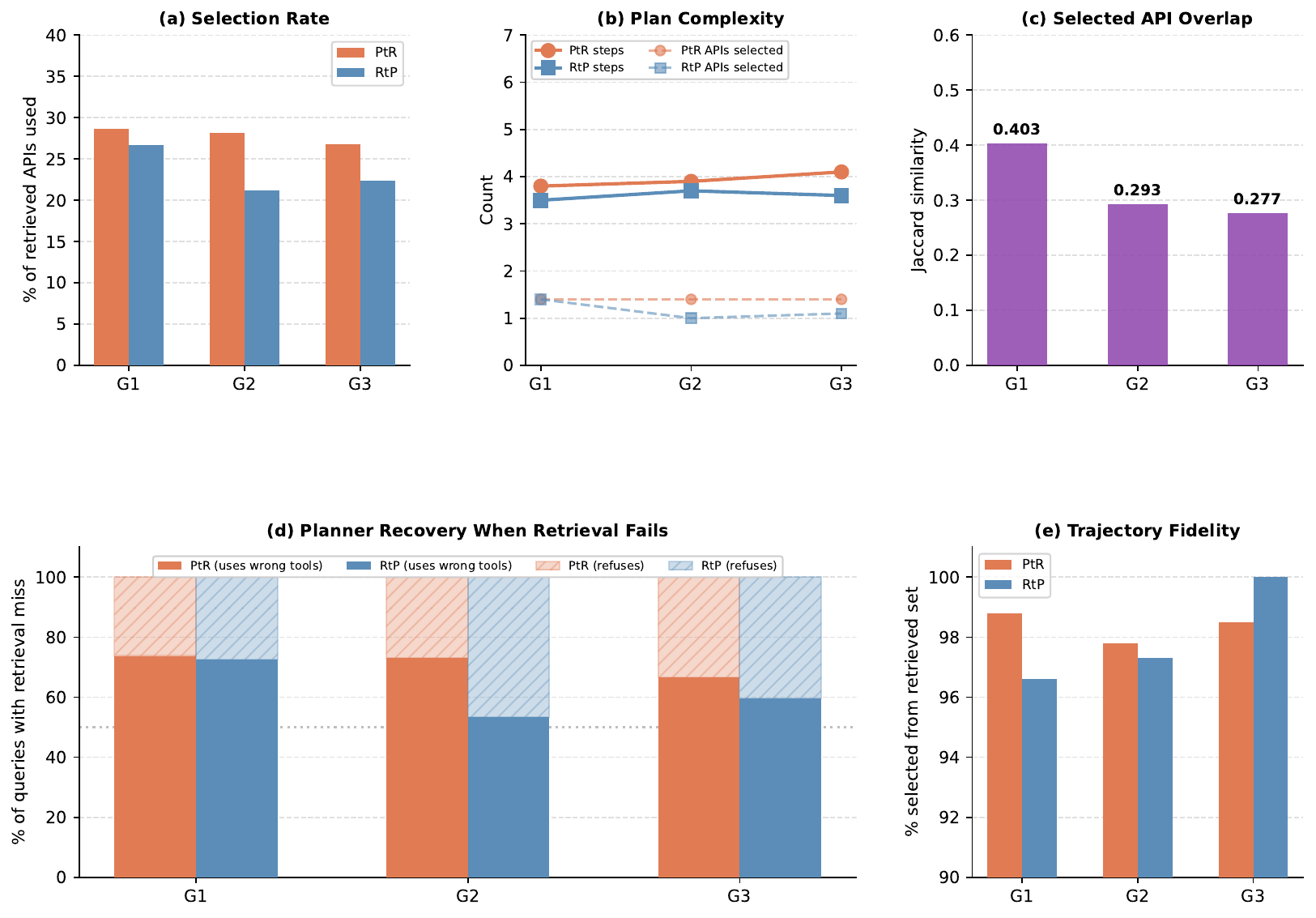}
\caption{Agent trajectory divergence across subsets: (a) selection rate = fraction of retrieved APIs the planner uses; (b) plan complexity = mean intended steps and APIs named; (c) Jaccard overlap of planner-selected API sets across architectures; (d) planner behavior when retrieval completely misses all Ground-Truth APIs; (e) trajectory fidelity = fraction of planner selections drawn from the retrieved set.}
\label{fig:trajectory}
\end{figure*}

Trajectory fidelity is at least $96\%$ across all subsets: the planner almost always selects from the retrieved set, confirming retrieval is the binding constraint on what the agent can do. Figure~\ref{fig:trajectory} shows five dimensions of trajectory divergence. The Jaccard similarity of planner-selected APIs decreases from $0.40$ (G1) to $0.28$ (G3), meaning even at the planner-selection level the architectures produce increasingly different tool call sequences. PtR's planner uses a higher fraction of retrieved APIs ($27$--$29\%$ vs.\ $21$--$27\%$), suggesting its stub-guided retrieval produces a more coherent candidate set that the planner trusts. On G3, PtR generates more ambitious plans ($4.1$ intended steps, $1.4$ APIs named) while RtP produces sparser plans ($3.6$ steps, $1.1$ APIs named), suggesting PtR's multi-stub retrieval gives the planner richer material to work with. When retrieval completely misses, the planner uses wrong tools ${\sim}70\%$ of the time for both architectures; but RtP's planner refuses more often on G2 ($47\%$ vs.\ $27\%$), likely because RtP's misses are obviously unrelated while PtR's stubs retrieve plausible-looking but functionally wrong tools.

\section{Discussion}

\subsection{Why the Crossover Occurs}
The crossover reflects the two faces of Intent-Registry Coupling. On G1, coupling is high: the raw query often contains the API's name or brand directly, giving RtP a direct embedding match while a planning step only abstracts away this signal. On G2/G3, coupling is low: a single embedding cannot capture multi-faceted intent. G2 queries require a mean of $6.5$ Ground-Truth APIs vs.\ $5.2$ on G1; PtR generates $2.4$--$2.5$ stubs per query, each anchoring to a different functional sub-need and retrieving independently. PtR's G2 advantage grows with $k$ (from $+0.014$ at $k{=}1$ to $+0.066$ at $k{=}20$) because each stub contributes its own top-$k$ candidates.

A natural concern is whether PtR's G2 advantage is simply an artifact of broader search: with $m{\approx}2.4$ stubs each retrieving top-$k$, PtR explores ${\sim}2.4k$ candidates before returning $k$ to the planner, while RtP searches exactly $k$. The $k$-ablation (Table~\ref{tab:k_ablation}) directly addresses this: at budget-equivalent $k{=}10$ (matching PtR's effective search width at $k{=}5$), RtP G2 recall reaches $0.186$, nearly identical to PtR at $k{=}5$ ($0.184$ in the ablation; $0.188{\pm}0.007$ in the main experiment). However, this budget-equivalent RtP passes $k{=}10$ APIs to the planner, halving precision relative to PtR's $k{=}5$ output. The G1 finding is unaffected: even at $k{=}10$, RtP G1 recall ($0.453$) far exceeds PtR at $k{=}5$ ($0.279$). Together, these results confirm that the crossover reflects structural properties of query-registry coupling rather than search breadth alone.

\subsection{Stub Hallucination: The Mechanism}
PtR generates $79$--$99\%$ unique stub names across subsets (Table~\ref{tab:stub_stats}), confirming the LLM is hallucinating tool names from scratch rather than recalling real API names. The cosine similarity between correct and incorrect stub-API matches is virtually identical (e.g., G1: $0.874$ vs.\ $0.872$; similarly small gaps on G2 and G3), meaning the embedding space provides \emph{no signal} to distinguish good routing from bad. Only $40.6\%$ of G1 stubs and $24.8\%$ of G3 stubs match a relevant API at all. Despite this, G3 recall favors PtR because multi-stub \emph{coverage} compensates for individual stub mismatches: $2.5$ stubs per query span different functional domains, and finding the right tool for each sub-need does not require every stub to succeed. Table~\ref{tab:examples} shows representative examples of each mode.

\begin{table*}[t]
\centering
\footnotesize
\setlength{\tabcolsep}{4pt}
\caption{Representative queries illustrating the dual role of stub hallucination. \textbf{Top}: RtP achieves perfect recall when the query contains the API name directly; PtR's generic stub abstracts away this signal and retrieves the wrong API. \textbf{Bottom}: PtR achieves higher recall when the query describes a functional need; stubs decompose multi-faceted intent into targeted sub-queries that RtP's single embedding cannot match. ``Ground-Truth API'' is the ToolBench gold API for that query.}
\label{tab:examples}
\begin{tabular}{>{\raggedright\arraybackslash}p{4.4cm}>{\raggedright\arraybackslash}p{4.4cm}>{\raggedright\arraybackslash}p{4.2cm}cc}
\toprule
\textbf{Query} & \textbf{PtR Stub(s)} & \textbf{Ground-Truth API} & \textbf{PtR R} & \textbf{RtP R} \\
\midrule
\multicolumn{5}{l}{\textit{RtP wins: query contains the API name directly}} \\[2pt]
``latest news about crime in Germany'' & \texttt{news\_article\allowbreak\_search\_api} & German Police and Crime News Live & 0.00 & 1.00 \\[2pt]
``programming memes from Reddit'' & \texttt{reddit\_api\_client} & Programming Memes Reddit/Get All & 0.00 & 1.00 \\[2pt]
``electricity bill PLN analysis'' & \texttt{get\_postpaid\allowbreak\_pln\_bill\_details} & Cek ID PLN/Get Data Tagihan & 0.00 & 1.00 \\[2pt]
``Testing out Sharing w/ Rachael'' & \texttt{api\_specification\allowbreak\_retriever} & Testing out Sharing w/ Rachael & 0.00 & 1.00 \\
\midrule
\multicolumn{5}{l}{\textit{PtR wins: query describes a capability; stubs bridge the semantic gap}} \\[2pt]
``surprise proposal, romantic lines + Instagram'' & \texttt{text\_generation}\newline\texttt{instagram\allowbreak\_downloader} & Flirty words + Instagram Downloader & 1.00 & 0.50 \\[2pt]
``monitor status of video uploads'' & \texttt{video\_ingest\allowbreak\_job\_status}\newline\texttt{upload\_urls} & Fake Brightcove/Ingest + Temp URLs & 1.00 & 0.00 \\[2pt]
``horoscope app, all zodiac signs'' & \texttt{get\_all\_zodiac\allowbreak\_signs\_data} & Zodiac Sign API/Getallsign & 1.00 & 0.00 \\[2pt]
``London trip: hotels + IP + currency'' & \texttt{hotel\_search}\newline\texttt{ip\_forensics}\newline\texttt{currency} & IP Forensics + Currency APIs & 0.50 & 0.00 \\
\bottomrule
\end{tabular}
\end{table*}

\subsection{Discovery as the Binding Constraint}
\label{sec:trajectory}
Trajectory fidelity (at least $96\%$, Figure~\ref{fig:trajectory}e) confirms the planner almost always selects from the retrieved set, making retrieval the bottleneck. As a result, architecture choice at retrieval time fully determines downstream behavior: PtR's planner generates more ambitious plans on G3 ($4.1$ intended steps, $1.4$ APIs named vs.\ $3.6$ steps, $1.1$ APIs) and uses a higher fraction of retrieved APIs ($27$--$29\%$ vs.\ $21$--$27\%$), reflecting the broader, more coherent candidate sets produced by stub-guided retrieval.

PtR's discovery call adds ${\sim}75$--$104\%$ token overhead (Appendix~\ref{app:cost}, Table~\ref{tab:cost}); on G1 this hurts ($-0.072$ recall) while on G2 it pays off ($+0.042$ recall). Beyond token cost, this discovery call sits on the critical path: because discovery and retrieval are sequential, PtR adds one full LLM round-trip of latency before retrieval and planning begin, so the overhead is both a token cost and a wall-clock latency penalty. Complexity-adaptive routing (using RtP for G1, PtR for G2/G3) saves ${\sim}16\%$ of total inference tokens, or ${\sim}36\%$ of the PtR discovery-call overhead, with no retrieval quality loss relative to always selecting the better architecture per subset, and eliminates the extra round-trip entirely for G1-type queries. The crossover maps naturally onto single-tool vs.\ multi-tool intent; practical proxies computable before any LLM call include query length, clause count, named-entity density, and the presence of conjunctive sub-goals — all of which correlate with the G1/G2 split in ToolBench. Most directly, RtP's own top-1 cosine similarity is a zero-cost routing signal already produced by the pipeline: a high top-1 score indicates the query names an API directly (high Intent-Registry Coupling, G1-like) and can stay with RtP, whereas a low or diffuse score indicates abstract multi-step intent (G2/G3-like) that benefits from PtR's stub decomposition — so a simple threshold on this score routes queries with no extra model and no extra call. We leave empirical validation of such classifiers to future work; complexity-adaptive routing over retrieval strategies has established precedent in Adaptive-RAG~\citep{jeong2024adaptiverag}, which routes between no-retrieval, single-step, and iterative retrieval by query complexity. Stub calibration (preserving proper nouns rather than abstracting them) could further reduce G1 misrouting. For queries requiring 12+ APIs, iterative discovery~\citep{yao2022react,schick2023toolformer} may be necessary regardless of architecture.

\section{Conclusion}

We present a systematic head-to-head comparison of Retrieve-then-Plan and Plan-then-Retrieve at industrial scale (50k APIs). Three findings stand out. First, there is a clear complexity-dependent crossover: RtP wins on single-tool queries; PtR wins significantly on intra-category multi-tool queries (G2: recall $+0.045$, $p{<}0.001$, $n{=}200$) and directionally on cross-category queries (G3: recall $+0.012$, $n{=}100$, not significant individually; pass rate difference is significant). This crossover holds at every retrieval budget ($k{=}1$--$20$), every registry size (1k--47k APIs), and replicates on gpt-oss-120b, confirming the result is architectural rather than model-specific. Second, stub hallucination is a double-edged sword: it destroys high-coupling signals on simple queries but productively decomposes multi-faceted intent on complex ones. Third, the retrieval step is the binding constraint on agent behavior: planners faithfully execute whatever tools they are given (at least $96\%$ fidelity), so architecture choice at retrieval time largely determines the downstream trajectory. These findings motivate complexity-adaptive routing as a practical direction: directing single-tool queries to RtP and multi-tool queries to PtR reduces discovery-call overhead by ${\sim}36\%$ with no retrieval quality loss. A cross-model replication on gpt-oss-120b (Appendix~\ref{app:model_generalization}) confirms the crossover directions are consistent across model families.

\section*{Limitations}
\label{sec:limitations}

This work is scoped as a controlled architectural comparison: we isolate the retrieval paradigm as the single varying factor, holding the planner, embedding model, and benchmark fixed (Gemini 2.5 Flash, GTE-base, ToolBench). This design choice enables clean attribution; Appendix~\ref{app:model_generalization} provides an initial replication on gpt-oss-120b confirming the crossover directions hold across model families, though generalization to other registry types (e.g., enterprise-internal APIs with different naming conventions) remains to be validated. As noted in \S\ref{sec:method}, we evaluate against ToolBench gold trajectories as a retrieval proxy; trajectory fidelity supports this as a valid proxy for planner behavior, though parameter correctness and final answer quality are not assessed. PtR's discovery and retrieval calls are sequential, adding one full LLM round-trip of latency before the planner begins; complexity-adaptive routing eliminates this overhead for G1-type queries. Token cost (reported in Appendix~\ref{app:cost}) is a proxy for this serial overhead structure; the resulting wall-clock latency depends on the serving infrastructure.

\section*{Availability}

The two prompts (discovery stub generation and planner) and the FAISS index configuration are provided verbatim in Appendix~\ref{app:prompts} and Appendix~\ref{app:faiss} for self-contained reproducibility.

\section*{Acknowledgments}

We thank our publication coordinator Helen Song for her support in preparing this work for release.

\bibliographystyle{acl_natbib}
\bibliography{reference}

\clearpage
\beginappendix

\section{Full Retrieval Metrics Table}
\label{app:main_table}

Table~\ref{tab:overall_app} reports pass rate, retrieval precision, and retrieval recall for both architectures across all three subsets. Pass rate (fraction of queries where the planner's intended plan selects at least one API) is reported here as a supplementary indicator; the main text focuses on precision and recall as the primary retrieval metrics. RtP leads on all G1 metrics; PtR leads on all G2 and G3 metrics.

\begin{table*}[t]
\centering
\small
\caption{Retrieval metrics by subset (4 independent runs each). RtP precision and recall are deterministic (no run-to-run variance; reported without std). $p$-values: one-sample $t$-test for precision/recall (PtR's 4-run mean vs.\ deterministic RtP); two-sample $t$-test for pass rate. \textbf{Bold}: best mean per metric. $^{**}p{<}0.01$, $^{***}p{<}0.001$, n.s.\ = not significant.}
\label{tab:overall_app}
\begin{tabular}{llccccc}
\toprule
& & \multicolumn{2}{c}{\textbf{RtP}} & \multicolumn{2}{c}{\textbf{PtR}} & \\
\cmidrule(lr){3-4}\cmidrule(lr){5-6}
\textbf{Subset} & \textbf{Metric} & \textbf{Mean} & \textbf{$\pm$Std} & \textbf{Mean} & \textbf{$\pm$Std} & \textbf{$p$} \\
\midrule
\multirow{3}{*}{G1 ($n{=}200$)} & Pass Rate     & $\mathbf{0.884}$ & $0.015$ & $0.840$ & $0.024$ & $0.022^{*}$ \\
 & Precision & $\mathbf{0.274}$ & $-$     & $0.209$ & $0.008$ & ${<}0.001^{***}$ \\
 & Recall    & $\mathbf{0.372}$ & $-$     & $0.289$ & $0.010$ & $0.002^{**}$ \\
\midrule
\multirow{3}{*}{G2 ($n{=}200$)} & Pass Rate     & $0.747$ & $0.015$ & $\mathbf{0.855}$ & $0.020$ & ${<}0.001^{***}$ \\
 & Precision & $0.167$ & $-$     & $\mathbf{0.204}$ & $0.007$ & $0.039^{*}$ \\
 & Recall    & $0.143$ & $-$     & $\mathbf{0.188}$ & $0.007$ & ${<}0.001^{***}$ \\
\midrule
\multirow{3}{*}{G3 ($n{=}100$)} & Pass Rate     & $0.732$ & $0.010$ & $\mathbf{0.820}$ & $0.014$ & ${<}0.001^{***}$ \\
 & Precision & $0.135$ & $-$     & $\mathbf{0.144}$ & $0.007$ & $0.052$ n.s. \\
 & Recall    & $0.160$ & $-$     & $\mathbf{0.172}$ & $0.008$ & $0.080$ n.s. \\
\bottomrule
\end{tabular}
\end{table*}

\section{Stub Generation Statistics}
\label{app:stubs}

Table~\ref{tab:stub_stats} reports stub generation statistics per subset. The high uniqueness rate ($79$--$99\%$) confirms stubs are hallucinated from scratch. The cosine similarity between stubs that correctly match a relevant API ($0.874$) and those that do not ($0.872$) is virtually identical, meaning the embedding space provides no signal to distinguish good routing from bad.

\begin{table*}[t]
\centering
\footnotesize
\setlength{\tabcolsep}{5pt}
\caption{Stub generation statistics by subset. \textit{Gen.\%}: fraction of queries generating stubs. \textit{Stubs/q}: mean stubs per query. \textit{Unique\%}: fraction of stub names unique across queries. \textit{Match\%}: fraction of stubs whose top-1 matched API is in Ground-Truth. \textit{Cos}: mean cosine similarity of stub to its top-1 match (virtually identical for correct vs.\ incorrect matches).}
\label{tab:stub_stats}
\begin{tabular}{lccccc}
\toprule
\textbf{Subset} & \textbf{Gen.\%} & \textbf{Stubs/q} & \textbf{Unique\%} & \textbf{Match\%} & \textbf{Cos} \\
\midrule
G1 & 98.5\% & 2.1 & 98.6\% & 40.6\% & 0.874 (correct) / 0.872 (incorrect) \\
G2 & 99.5\% & 2.4 & 93.6\% & 38.3\% & 0.843 / 0.840 \\
G3 & 100\% & 2.5 & 79.1\% & 24.8\% & 0.858 / 0.854 \\
\bottomrule
\end{tabular}
\end{table*}

Table~\ref{tab:search_width} reports the effective search width for each architecture and subset at $k{=}5$. At equal nominal $k$, PtR explores ${\sim}2$--$2.5\times$ more candidates than RtP before returning the top-$k$ to the planner.

\begin{table*}[htb]
\centering
\small
\caption{Effective search width at $k{=}5$. RtP searches exactly $k$ candidates; PtR searches $k{\times}$Stubs/q (union, deduped) then returns top-$k$. Budget-equivalent RtP recall from Table~\ref{tab:k_ablation}.}
\label{tab:search_width}
\begin{tabular}{lccccc}
\toprule
\textbf{Subset} & \textbf{Stubs/q} & \textbf{PtR eff.\ width} & \textbf{RtP eff.\ width} & \textbf{Budget-equiv.\ $k$} & \textbf{RtP recall} \\
\midrule
G1 & 2.1 & ${\approx}10$ & 5 & 10 & 0.453 \\
G2 & 2.4 & ${\approx}12$ & 5 & 10 & 0.186 \\
G3 & 2.5 & ${\approx}13$ & 5 & 10 & 0.217 \\
\bottomrule
\end{tabular}
\end{table*}

\section{Trajectory Analysis Detail}
\label{app:trajectory}

Table~\ref{tab:traj} reports quantitative trajectory divergence metrics across subsets. All metrics are computed over paired queries where both architectures produced a scorable response.

\begin{table}[htb]
\centering
\small
\caption{Trajectory divergence metrics. Selection rate = fraction of retrieved APIs the planner uses. Jaccard = similarity of planner-selected API sets. Fidelity = fraction of planner selections drawn from the retrieved set.}
\label{tab:traj}
\begin{tabular}{lcccc}
\toprule
\textbf{Metric} & \textbf{Arch.} & \textbf{G1} & \textbf{G2} & \textbf{G3} \\
\midrule
\multirow{2}{*}{Selection rate} & PtR & 28.6\% & 28.2\% & 26.8\% \\
 & RtP & 26.7\% & 21.2\% & 22.4\% \\
\midrule
Jaccard (selected) & both & 0.40 & 0.29 & 0.28 \\
\midrule
\multirow{2}{*}{Plan steps} & PtR & 3.8 & 3.9 & 4.1 \\
 & RtP & 3.5 & 3.7 & 3.6 \\
\midrule
\multirow{2}{*}{Fidelity} & PtR & 98.8\% & 97.8\% & 98.5\% \\
 & RtP & 96.6\% & 97.3\% & 100\% \\
\midrule
Refusal on miss & PtR & 26\% & 27\% & 33\% \\
 & RtP & 28\% & 47\% & 40\% \\
\bottomrule
\end{tabular}
\end{table}

\section{Tool Selection Divergence}
\label{app:divergence}

Table~\ref{tab:divergence_app} shows tool selection divergence between RtP and PtR across all subsets. Divergence increases monotonically with query complexity, reaching $100\%$ on G3, where the two architectures retrieve from entirely non-overlapping API sets for every query.

\begin{table}[htb]
\centering
\small
\caption{Tool selection divergence: fraction of queries where the two architectures retrieve different top-$k$ API sets.}
\label{tab:divergence_app}
\begin{tabular}{lccc}
\toprule
\textbf{Subset} & \textbf{$n$} & \textbf{Identical} & \textbf{Divergent} \\
\midrule
G1 & 200 & 2.5\% & 97.5\% \\
G2 & 200 & 1.0\% & 99.0\% \\
G3 & 100 & 0.0\% & 100.0\% \\
\bottomrule
\end{tabular}
\end{table}

When tools diverge, the recall-win pattern flips: RtP wins on 63 G1 queries vs.\ PtR's 33 (mean recall $0.369$ vs.\ $0.294$), but PtR wins on 69 G2 queries vs.\ RtP's 30 (mean recall $0.184$ vs.\ $0.142$).

\section{Inference Cost Analysis}
\label{app:cost}

Table~\ref{tab:cost} quantifies PtR's token overhead against its recall benefit. PtR's discovery call adds ${\sim}175$--$200$ output tokens per query. On G1 this overhead hurts (higher cost, lower recall). On G2 it is well justified ($+0.042$ recall for $+80\%$ tokens).

\begin{table}[htb]
\centering
\small
\caption{Inference cost and recall benefit of PtR over RtP. The sign of $\Delta$R matches the crossover in Table~\ref{tab:overall_app}, where the G1 and G2 recall differences are statistically significant ($p{<}0.01$) and G3 is not.}
\label{tab:cost}
\begin{tabular}{lcccc}
\toprule
& \textbf{RtP} & \textbf{PtR} & \textbf{Overhead} & \textbf{$\Delta$R} \\
\midrule
G1 & ${\sim}226$ & ${\sim}396$ & $+75\%$ & $-0.072$ \\
G2 & ${\sim}238$ & ${\sim}429$ & $+80\%$ & $+0.042$ \\
G3 & ${\sim}221$ & ${\sim}450$ & $+104\%$ & $+0.017$ \\
\bottomrule
\end{tabular}
\end{table}

\section{Stub-Hallucination Examples}
\label{app:halluc}

Table~\ref{tab:halluc} shows G1 queries where PtR achieves zero recall (stub abstracts away the brand/product name) while RtP achieves perfect recall via direct embedding match.

\begin{table*}[t]
\centering
\small
\caption{Stub-hallucination examples: PtR's over-generic stubs miss niche APIs that RtP matches directly.}
\label{tab:halluc}
\begin{tabular}{>{\raggedright\arraybackslash}p{4.2cm}>{\raggedright\arraybackslash}p{3.6cm}>{\raggedright\arraybackslash}p{6.2cm}}
\toprule
\textbf{Query} & \textbf{PtR Stub} & \textbf{RtP Retrieved (correct)} \\
\midrule
``crime news in Germany'' & \texttt{news\_search\_api} & German Police and Crime News Live \\
``programming memes, Reddit'' & \texttt{reddit\_api\_client} & Programming Memes Reddit/Get All Memes \\
``PLN electricity bill'' & \texttt{get\_pln\allowbreak\_bill\_details} & Cek ID PLN/Get Data Tagihan \\
``Testing out Sharing w/ Rachael'' & \texttt{api\_spec\allowbreak\_retriever} & Testing out Sharing w/ Rachael/Input test \\
\bottomrule
\end{tabular}
\end{table*}

\section{PtR Win Examples (G1, G2, G3)}
\label{app:ptr_wins}

\paragraph{G1 wins (functional-need queries).}
\begin{itemize}
\item ``\textit{monitor status of video uploads}'' $\to$ stubs \texttt{get\_video\_ingest\_job\_status} + \texttt{generate\_temporary\_upload\_urls} retrieve both correct Fake Brightcove APIs (recall 1.0); RtP retrieves unrelated video encoding tools (recall 0.0).
\item ``\textit{company event in Dallas, need airports}'' $\to$ stub \texttt{airport\_information\_api} retrieves \texttt{Airports Finder/Get Airports By City} (recall 1.0); RtP drifts to \texttt{Airport Data/Airports By City} (recall 0.0).
\item ``\textit{horoscope app, all zodiac signs}'' $\to$ stub \texttt{get\_all\_zodiac\_signs\_data} retrieves \texttt{Zodiac Sign API/Getallsign} (recall 1.0); RtP retrieves unrelated astrology APIs (recall 0.0).
\end{itemize}

\paragraph{G2 wins (multi-stub decomposition).}
\begin{itemize}
\item ``\textit{surprise proposal, romantic lines + Instagram media}'' $\to$ PtR decomposes into \texttt{text\_generation\allowbreak\_tool} + \texttt{instagram\_media\allowbreak\_downloader\_tool}, achieving recall 1.0; RtP achieves 0.5.
\item ``\textit{London trip: hotels + IP forensics + currency}'' $\to$ 3 stubs span 3 functional areas, achieving recall 0.5; RtP achieves 0.0 (single embedding drifts to hotel listings).
\end{itemize}

\paragraph{G3 wins (cross-category decomposition).}
\begin{itemize}
\item ``\textit{share news article, create shortened URL}'' $\to$ PtR generates \texttt{url\_shortener} + \texttt{slashtag\_suggestion} + \texttt{domain\_lister}, covering 3 tools in the URL shortening toolchain (recall 0.33); RtP retrieves generic news APIs (recall 0.0).
\end{itemize}

\section{G2/G3 RtP Win Examples}
\label{app:rtp_wins}

\paragraph{G2 -- RtP wins when stubs over-generalize.}
\begin{itemize}
\item ``\textit{trip to Greece, explore energy market news}'' $\to$ PtR generates \texttt{general\_news\_search} + \texttt{eurosport\_football\_news} (recall 0.0); RtP matches ``Greek'' and ``Energy'' keywords, retrieving \texttt{Greek News in English} + \texttt{Energy Price News} (recall 0.5).
\item ``\textit{team-building event: trivia + gaming}'' $\to$ PtR stubs miss specific API brands (recall 0.0); RtP retrieves \texttt{Trivia by API-Ninjas} + \texttt{Epic Free Games} directly (recall 0.4).
\end{itemize}

\paragraph{G3 -- RtP wins when queries contain niche brand references.}
\begin{itemize}
\item ``\textit{NYC trip, local craft breweries}'' $\to$ PtR generates \texttt{local\_business\_search} + \texttt{recipe\_search} (wrong decomposition, recall 0.0); RtP matches ``brewery'' directly, retrieving the correct Brewery API (recall 0.5).
\item ``\textit{birthday party, create video for friend}'' $\to$ PtR stub \texttt{vimeo\_video\_search} matches \texttt{searchUsers} instead of \texttt{searchVideos} (recall 0.0); RtP retrieves \texttt{Vimeo/searchVideos} via raw query (recall 0.25).
\end{itemize}

\section{Scaling Study Details}
\label{app:scaling}

Sub-registries are constructed by including all Ground-Truth APIs for the subset under study plus proportionally sampled distractor APIs from all 50 ToolBench categories to reach the target size; the G1 and G2 recall curves use their respective Ground-Truth sets, and the divergence column is measured on the G1 sub-registries. Each size is run with 5 random seeds (different distractor samples); the full registry (46{,}984 APIs) has no randomness and is run once. Because G2's Ground-Truth set (1{,}048 APIs) already exceeds the 1k target, the smallest G2 point is gold-only, slightly larger than 1k, and has zero seed variance. PtR reuses the saved stubs from the main experiment (stub generation is independent of registry size), so no additional LLM calls are made; only FAISS retrieval is recomputed per sub-registry. Table~\ref{tab:scaling_raw} reports the full numerical results.

\begin{table*}[t]
\centering
\small
\caption{Scaling study: RtP vs.\ PtR recall by registry size for the G1 (RtP-advantage) and G2 (PtR-advantage) subsets, with G1 tool-selection divergence ($\pm$ std across 5 random seeds; full registry has no randomness). \textbf{Bold} marks the better architecture per subset: the crossover (RtP $>$ PtR on G1, PtR $>$ RtP on G2) holds at every registry size. The full-registry recall gaps are statistically significant on both subsets (Table~\ref{tab:overall_app}). G2's Ground-Truth set (1{,}048 APIs) exceeds the 1k target, so the smallest G2 point is gold-only (no distractors, zero variance).}
\label{tab:scaling_raw}
\begin{tabular}{rccccc}
\toprule
& \multicolumn{2}{c}{\textbf{G1 Recall}} & \multicolumn{2}{c}{\textbf{G2 Recall}} & \\
\cmidrule(lr){2-3}\cmidrule(lr){4-5}
\textbf{Registry size} & \textbf{RtP} & \textbf{PtR} & \textbf{RtP} & \textbf{PtR} & \textbf{Div\%} \\
\midrule
1{,}000  & $\mathbf{.646}{\pm}.001$ & $.586{\pm}.004$ & $.339{\pm}.000$ & $\mathbf{.366}{\pm}.000$ & $88.4{\pm}0.0$ \\
2{,}000  & $\mathbf{.603}{\pm}.004$ & $.544{\pm}.009$ & $.300{\pm}.006$ & $\mathbf{.336}{\pm}.005$ & $89.2{\pm}1.2$ \\
5{,}000  & $\mathbf{.536}{\pm}.008$ & $.470{\pm}.012$ & $.249{\pm}.003$ & $\mathbf{.294}{\pm}.004$ & $92.5{\pm}1.1$ \\
10{,}000 & $\mathbf{.484}{\pm}.002$ & $.414{\pm}.008$ & $.215{\pm}.005$ & $\mathbf{.260}{\pm}.003$ & $94.1{\pm}1.2$ \\
20{,}000 & $\mathbf{.430}{\pm}.002$ & $.357{\pm}.003$ & $.184{\pm}.003$ & $\mathbf{.226}{\pm}.003$ & $95.6{\pm}0.8$ \\
46{,}984 & $\mathbf{.375}$ & $.297$ & $.143$ & $\mathbf{.185}$ & $97.5$ \\
\bottomrule
\end{tabular}
\end{table*}

\section{Model Generalization: gpt-oss-120b Replication}
\label{app:model_generalization}

We replicate the main experiment using gpt-oss-120b~\citep{openai2025gptoss120bgptoss20bmodel}, holding all other components fixed (GTE-base embeddings, same FAISS index, same ToolBench test sets, $k{=}5$), to test whether the findings from the main text hold for a different model family. Table~\ref{tab:model_generalization} reports results.

\begin{table*}[t]
\centering
\small
\caption{gpt-oss-120b replication (mean$\,\pm\,$std, 4 runs). RtP precision and recall are deterministic. \textbf{Bold}: better architecture (by mean). Superscripts on PtR values denote significance of the PtR--RtP difference (one-sample $t$-test vs.\ deterministic RtP for precision/recall; two-sample for pass rate): $^{*}p{<}0.05$, $^{**}p{<}0.01$, $^{***}p{<}0.001$, n.s.\ = not significant. The G1 (RtP) and G2 (PtR) crossover replicates and is significant on precision and recall; on G3 the differences are small and the recall gap weakly reverses toward RtP, indicating G3 behavior is model-dependent.}
\label{tab:model_generalization}
\begin{tabular}{lcccccc}
\toprule
& \multicolumn{2}{c}{\textbf{Precision}} & \multicolumn{2}{c}{\textbf{Recall}} & \multicolumn{2}{c}{\textbf{Pass Rate}} \\
\cmidrule(lr){2-3}\cmidrule(lr){4-5}\cmidrule(lr){6-7}
\textbf{Subset} & RtP & PtR & RtP & PtR & RtP & PtR \\
\midrule
G1 & \textbf{0.274} & $0.186{\pm}0.006^{***}$ & \textbf{0.372} & $0.259{\pm}0.012^{***}$ & $\mathbf{0.855{\pm}0.018}$ & $0.786{\pm}0.024^{**}$ \\
G2 & 0.168 & $\mathbf{0.207{\pm}0.004}^{***}$ & 0.143 & $\mathbf{0.194{\pm}0.003}^{***}$ & $0.826{\pm}0.029$ & $0.820{\pm}0.029^{\text{n.s.}}$ \\
G3 & \textbf{0.135} & $0.123{\pm}0.008^{\text{n.s.}}$ & \textbf{0.160} & $0.146{\pm}0.008^{*}$ & $0.790{\pm}0.008$ & $\mathbf{0.805{\pm}0.013}^{\text{n.s.}}$ \\
\bottomrule
\end{tabular}
\end{table*}

The main findings replicate. On G1, RtP outperforms PtR on precision ($+0.088$), recall ($+0.113$), and pass rate ($+0.069$). On G2, PtR outperforms RtP on precision ($+0.039$) and recall ($+0.051$); pass rate is a statistical tie ($0.826{\pm}0.029$ vs.\ $0.820{\pm}0.029$). On G3, RtP has higher precision ($0.135$ vs.\ $0.123{\pm}0.008$, n.s.) and recall ($0.160$ vs.\ $0.146{\pm}0.008$, $p{=}0.035$) while PtR leads on pass rate ($0.805{\pm}0.013$ vs.\ $0.790{\pm}0.008$, n.s.); unlike Gemini, the small G3 recall gap reverses toward RtP here, underscoring that G3 (the smallest set) is the least stable across model families. RtP retrieval metrics are identical to the Gemini results, confirming they depend only on the query embedding and registry. The slightly lower PtR precision and recall for gpt-oss-120b vs.\ Gemini on G1 suggests stub quality varies across model families, though the crossover direction is preserved.

\section{Prompts}
\label{app:prompts}

We use two zero-shot system prompts, reproduced verbatim below. The discovery prompt (PtR only) elicits abstract tool stubs from the query alone; the planner prompt (both architectures) selects APIs from the retrieved set.

\begin{lstlisting}[title={\textbf{Discovery prompt (stub generation)}}]
You are a tool discovery assistant. Given a user query, identify what tools or APIs would be needed to answer it. You do NOT have access to any tool registry — instead, describe the tools you think are needed based on the query alone.

Respond ONLY with a JSON array:
[
  {"tool_name": "hypothetical_tool_name", "reasoning": "why this tool is needed"},
  ...
]

Be specific about tool functionality. Use snake_case for tool names.
\end{lstlisting}

\begin{lstlisting}[title={\textbf{Planner prompt}}]
You are a tool-augmented assistant. Given a user query and a set of available APIs, select the best API(s) to answer the query and produce a step-by-step plan.

Respond in JSON:
{
  "selected_apis": ["tool_name/api_name", ...],
  "plan": "step-by-step plan",
  "answer": "final answer to the user"
}
\end{lstlisting}

\section{FAISS Index Configuration}
\label{app:faiss}

Each API is embedded with \texttt{thenlper/gte-base} (768-dimensional, L2-normalized) from the string \texttt{"\{tool\_name\}/\{api\_name\}: \{description\}"}, and all APIs are indexed with \texttt{faiss.IndexFlatIP} (exact inner-product search, equivalent to cosine similarity for normalized vectors).

\applefootnote{ \textcolor{textgray}{\sffamily Apple and the Apple logo are trademarks of Apple Inc., registered in the U.S. and other countries and regions.}}

\end{document}

%% file: apple_preamble.tex
\usepackage{amsmath}
\usepackage{enumerate}
\usepackage{algorithm}
\usepackage{algpseudocode}
\usepackage{amsfonts}
\usepackage{amsthm}
\usepackage{cleveref}
\usepackage{diagbox}
\usepackage{colortbl}
\usepackage{amssymb}
\usepackage{xspace}
\usepackage{wrapfig}
\usepackage{adjustbox}
\usepackage{tabularx}
\usepackage{booktabs}
\usepackage{mathtools}
\usepackage{tikz}
\usepackage{enumitem}
\usepackage{silence}
\usepackage{dsfont}
\usepackage[table]{xcolor}
\usepackage[dvipsnames]{xcolor}
\usepackage{multirow}
\usepackage{makecell}
\usepackage{xfakebold}
\input{math_commands}

\definecolor{textgray}{HTML}{6E6E73}
\usetikzlibrary{positioning, calc}
\usetikzlibrary{decorations.pathmorphing}

\makeatletter
\patchcmd{\wrong@fontshape}{\@gobbletwo}{}{}{}
\makeatother
\numberwithin{equation}{section}
\makeatletter
\AtBeginDocument{
  \urlstyle{sf}
  
}
\makeatother

\definecolor{light}{RGB}{125, 125, 125}
\crefname{tcb@cnt@pbox}{code}{code}
\Crefname{tcb@cnt@pbox}{Code}{Code}
\crefname{assumption}{assumption}{assumption}
\Crefname{assumption}{Assumption}{Assumptions}

\newtcolorbox[auto counter]{pbox}[2][]{
  colback=white,
  title=Code~\thetcbcounter: #2,
  #1,fonttitle=\sffamily,
  fontupper=\sffamily,
  arc=2pt,
  colframe=bgcolor,
  coltitle=fgcolor,
  colbacktitle=bgcolor,
  toptitle=0.25cm,
  bottomtitle=0.125cm
}

\makeatletter
\newcommand\applefootnote[1]{%
  \begingroup
  \renewcommand\thefootnote{}%
  \renewcommand\@makefntext[1]{\noindent##1}%
  \footnote{#1}%
  \addtocounter{footnote}{-1}%
  \endgroup
}
\makeatother

\definecolor{cverbbg}{gray}{0.90}

%% file: math_commands.tex
\usepackage{amsmath,amsfonts,bm}

\def\eqref#1{equation~\ref{#1}}

\def\1{\bm{1}}

\DeclareMathAlphabet{\mathsfit}{\encodingdefault}{\sfdefault}{m}{sl}
\SetMathAlphabet{\mathsfit}{bold}{\encodingdefault}{\sfdefault}{bx}{n}

